\documentclass[letterpaper, 10 pt, conference]{ieeeconf}  

\IEEEoverridecommandlockouts                              

\usepackage{amsmath} 
\usepackage{amssymb}  

\usepackage{graphicx}

\usepackage{booktabs}
\usepackage{multirow}

\usepackage{xcolor}  
\usepackage{cite} 
\usepackage{lineno,hyperref}
\usepackage[capitalise]{cleveref} 

\newcommand{\LinkI}{\textit{Ego-observation}\,\,$\rightarrow$\,\,\textit{Ego-behavior}}
\newcommand{\LinkII}{\textit{Ego-behavior}\,\,$\rightarrow$\,\,\textit{Collective-behavior}}
\newcommand{\LinkIII}{\textit{Ego-observation}\,\,$\rightarrow$\,\,\textit{Ego-behavior}\,\,$\rightarrow$\,\,\textit{Collective-behavior}}

\title{Unveiling Complex Collective Behaviors from Simple Rewards}

\author{Yize Mi$^{1,2}$, Jianan Li$^{3,2}$, Liang Li$^{4,5,6,7}$, and Shiyu Zhao$^{2*}$
	\thanks{*Corresponding author}%
	\thanks{$^{1}$College of Computer Science and Technology, Zhejiang University, Hangzhou, China.}%
	\thanks{$^{2}$WINDY Lab, School of Engineering, Westlake University, Hangzhou, China. {\small \{miyize, lijianan, zhaoshiyu\}@westlake.edu.cn}}%
	\thanks{$^{3}$Shanghai AI Laboratory, 200030, Shanghai, China}%
	\thanks{$^{4}$Department of Collective Behaviour, Max Planck Institute of Animal Behavior, Konstanz, Germany. lli@@ab.mpg.de}%
	\thanks{$^{5}$Department of Computer and Information Science, University of Konstanz, Konstanz, Germany.}%
	\thanks{$^{6}$Centre for the Advanced Study of Collective Behaviour, University of Konstanz, Konstanz, Germany.}%
	\thanks{$^{7}$Department of Biology, University of Konstanz, Konstanz, Germany.}%
	\thanks{This work was supported by the Brain Science and Brain-like Intelligence Technology — National Science and Technology Major Project (Grant No. 2022ZD0208800)}%
}

\begin{document}

\maketitle
\thispagestyle{empty}
\pagestyle{empty}

\begin{abstract}
Multi-agent Reinforcement Learning (MARL) holds great potential for robot swarms, but the black-box nature of neural policies complicates strategic analysis, limiting multi-robot applications. Furthermore, complex swarm behaviors can surprisingly emerge from simple rewards without explicit aggregation incentives. Unveiling the mechanisms behind this emergence is critical, but the disconnection between simple rewards and collective behaviors exacerbates interpretability challenges. This paper aims to reveal the hidden mechanisms in this process. We propose a two-stage EEC (\LinkIII) explanatory framework. This includes a novel analytical tool called the Agent Response Map (ARM), which reveals agents' decision-making patterns across space and identifies regions of aggregation and avoidance.  ARM reveals that the robots implicitly learn the geometric fields of the environment and utilize these structures as desired targets for coordinated movement. We validate this finding across two distinct tasks: a cooperative multi-robot shape assembly and a competitive predator-prey pursuit-evasion. 1) In the cooperative task, ARM identifies the unoccupied target interior as the desired destination for robot navigation. As the center becomes occupied, this target region automatically shifts toward the boundary, demonstrating the robots' capacity to autonomously explore unoccupied areas. 2) In the competitive task, ARM surprisingly identifies the boundary of the predators' Voronoi diagram as the convergence destination for prey agents. Together, these two tasks demonstrate the capability of ARM to discover the hidden geometric structures underlying MARL policies in robot swarms. We perform quantitative analysis to validate the predictive accuracy of ARM.  The insights and tools presented in this paper may provide a new perspective on bridging the interpretability gap in applying black-box MARL to multi-robot systems.
\end{abstract}
\IEEEpeerreviewmaketitle

\section{Introduction}
Collective behaviors (e.g., starling flocks, fish schools, and sheep herds) are complex and fascinating phenomenon in nature \cite{sumpter_collective_2010}, attracting widespread attention across various fields. A prevailing view is that survival pressure plays a central role in driving animals to form groups, as collective behavior can increase individual survival chances in the face of predation or environmental risks \cite{beauchamp2013social}. Yet rigorously quantifying and explaining such phenomena remains challenging. Various
traditional models have been proposed to explain these collective behaviors, constructed from hand-designed rules \cite{vicsek_novel_1995, reynolds1987flocks, couzin_collective_2002}. While these models can replicate certain emergent patterns, their simplified interaction mechanisms fall short of capturing the complexity of real biological systems.

Recently, reinforcement learning (RL) has been used to simulate adaptive strategies in biological organisms \cite{durve_learning_2020,monter_dynamics_2023}, with multi-agent reinforcement learning (MARL) enabling simulations of complex agent interactions and optimizing collective behavior via rewards. For example, \cite{durve_learning_2020} penalizes agents that ``lose neighbors", while \cite{monter_dynamics_2023} encourages prey to gather by reducing dangerous areas. However, these methods often rely on explicit hand-crafted rewards and therefore provide limited fundamental insights into why natural organisms would follow such designed rules.

In contrast, a recent study \cite{li_predatorprey_2023} adopts a sole survival-pressure reward function and removes other hand-crafted reward terms within a predator-prey coevolution framework. It demonstrates the natural emergence of collective behaviors under simple incentives. This naturally raises the question: how can complex collective behaviors emerge from such simple rewards? 

Clarifying this mechanism not only provides a mechanistic account of emergent swarm intelligence but also facilitates the deployment of swarm-control policies in multi-robot applications. However, existing interpretability work \cite{bush2025interpreting, delfosse2024interpretable} struggles to systematically explain why local behavior at the individual level can give rise to swarm intelligence.

\begin{figure*}[h]
	\centering 
	\includegraphics[width=1\linewidth]{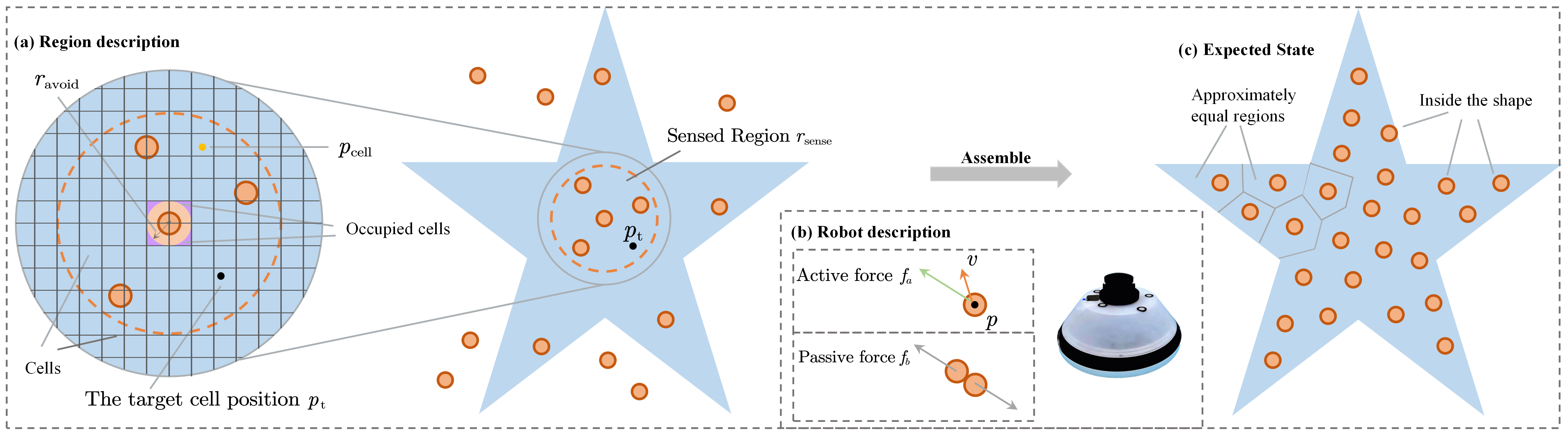}
	\caption{Shape assembly environment description. Agents are required to navigate to designated regions and self-organize into a user-specified target shape. (a) The target region description. (b) The robots description. (c) The desired assembly state.}\label{fig:env_shape_assembly}
\end{figure*}

To address this question, we propose a two-stage EEC (\LinkIII) framework to trace how egocentric inputs drive individual actions and how those actions scale to swarm patterns: 1) \LinkI, where key influential observation features are identified through SHAP (SHapley Additive exPlanations) \cite{lundberg_unified_2017} to quantify their contributions to the learned policy, followed by ablation experiments to validate the causal influence. This allows us to isolate the features most critical to the RL network's decision-making; 2) \LinkII, where we introduce the Agent Response Map (ARM), a novel explanation tool that reveals agents' decisions under different observations, and explains how ego-behavior yields swarm behavior by exposing common geometric fields across observations.

The proposed explanation method is demonstrated by two distinct tasks: a cooperative multi-robot shape assembly and a competitive predator-prey pursuit-evasion. Across these two distinct tasks, we surprisingly discover that ARM reveals a commonality in robot collective behaviors: robots implicitly learn latent risk fields and geometric invariants of the environment, exploiting these structures as desired targets for coordinated motion. Thus, the ARM method provides insight into why simple rewards can give rise to complex collective behavior.

To our knowledge, ARM is the first tool to reveal how agents coordinate their actions to induce collective behaviors, while existing interpretability methods focus on feature attribution or trajectory summarization. Therefore, no direct baseline exists. Instead, we validate ARM using two qualitative criteria: 1) In the competitive case, we compare our RL-based policy against three typical swarm models \cite{vicsek_novel_1995, reynolds1987flocks, couzin_collective_2002} to demonstrate the convergence of prey agents. 2) The cooperative case is verified by comparing the variation in critic values for robots inside versus outside the target region.

These results suggest that understanding the mechanisms behind emergent collective behavior can provide new insights for MARL and inform principled design of robust swarm-control policies.

\section{Related Work}
\subsection{Interpretable Reinforcement Learning}

Interpretability research in reinforcement learning (RL) can be broadly classified into three categories \cite{milani_explainable_2024}: Feature Importance (FI), Learning Process (LP), and Policy-Level (PL) analysis.

1) FI methods focus on explaining agent actions by identifying the observation features that most influence decision-making at each time step. Common techniques include SHAP values \cite{lundberg_unified_2017}, Integrated Gradients \cite{sundararajan2017axiomatic}, and Saliency Maps \cite{atrey_exploratory_2019}. Although FI methods have been applied to analyze agent behaviors in MARL \cite{heuillet2022collective}, they are typically limited to single-step decisions and struggle to capture long-term behavioral patterns. {2) LP methods aim to understand how the training process shapes agent behavior by identifying influential training samples or critical states \cite{cheng_statemask_2023}. However, LP approaches generally do not explain action selection given a specific observation.} 3) PL methods summarize long-term behavior through abstraction or representative trajectories \cite{boggess_toward_2022}. 

However, the high computational complexity restricts them to small-scale single-agent environments, rendering them unsuitable for large-scale multi-agent scenarios. Furthermore, these methods primarily translate policies into rules without explaining the mechanistic origin of \textit{why} such behaviors evolve.

\subsection{Collective Behavior Generation via MARL}
Multi-Agent Reinforcement Learning (MARL) serves as a powerful framework for modeling natural collective behaviors. Studies have demonstrated emergent flocking and foraging \cite{hahn2019emergent}, coordinated herding \cite{ritz2021sustainable}, transhipment scheduling\cite{wang2026uav} and target-directed swimming \cite{tovey2024emergence}.  {Notably, \cite{heuthe2024counterfactual} recently applied MARL to control 200 laser-actuated microrobots. By introducing a counterfactual reward scheme, the swarm learned to cooperatively transport cargo, mirroring ant colony dynamics.}

{However, these approaches typically rely on reward functions that explicitly encourage clustering or coordination. Such hand-crafted incentives risk embedding human heuristics, potentially diverging from the fundamental mechanisms driving natural swarm emergence.}

\section{Problem Setup}\label{sec:env}
\subsection{Cooperative Multi-robot Shape Assembly}
First, we consider a cooperative multi-robot task: shape assembly. This task represents a longstanding benchmark in cooperative control since it requires that robots satisfy a global target shape constraint while simultaneously maintaining spacing, avoiding collisions, and preserving a uniform formation.

\textit{\textbf{Region Description: }}The multi-robot shape assembly problem is formulated as a two-stage process: 1) A target geometric configuration is first defined, as depicted in \cref{fig:env_shape_assembly}(a); 2) Robots depicted in \cref{fig:env_shape_assembly}(b) use this shape as a goal and, through continuous local interactions with neighbors during motion, uniformly fill the shape, shown in \cref{fig:env_shape_assembly}(c). This connected target domain is spatially discretized into a grid consisting of $n_{\text{cell}}$ units. Each grid cell is defined by a side length of $l_{\text{cell}}$, with its geometric center located at $p_{\text{cell}}$. The position of a specific target cell is designated as $p_{\text{t}}$.

\textit{\textbf{Robot Description: }}The state of the $i$-th robot is characterized by a 2D vector ${{p}}_i\in\mathbb{R}^2$. The robot's motion is driven by the active and
passive forces.  ${f}_a\in\mathbb{R}^2$ represents the control input generated by the actor network, while ${f}_b\in\mathbb{R}^2$ is an elastic force following Hooke's Law. We define $r_{\text{avoid}}$ as the collision radius. Physical collisions are registered when the inter-agent distance falls below $2r_{\text{avoid}}$, whereas a grid cell is deemed occupied if a robot approaches within $r_{\text{avoid}}$ of the cell center.  The robot's dynamic model is $\dot{{p}}_i={v}_i, \quad \dot{{v}}_i=\left({f}_{\mathrm{a}}+{f}_{\mathrm{b}}\right) / m_i$, where $m_i$ denotes mass. This formulation aligns with our holonomic omnidirectional mobile platforms, as illustrated in \cref{fig:env_shape_assembly}(b).

\textit{\textbf{Observation and Reward Design: }}Each agent utilizes a four-component observation vector comprising: 1) its own state; 2) the relative states of neighboring agents; 3) the relative position  to the target cell; and 4) the relative positions of observed, unoccupied cells within the sensing range $r_{\text{sense}}$. We define $n_{\text{hn}}$ and $n_{\text{hc}}$ as the maximum capacity for tracking neighbors and cells, respectively, utilizing zero-padding for sparse scenarios. Consequently, the total observation dimensionality is $(6+4n_{\text{hn}}+2n_{\text{hc}})$.

The reward function design is considered to satisfy three conditions.  If all conditions are met,
the reward is 1; otherwise, it is 0: (1) spatial containment within the target shape; (2) successful collision avoidance; and (3) effective exploration of empty regions, formalized as:  $\left|\sum_{k \in \mathcal{C}_i} \rho_k {p}_k / \sum_{k \in \mathcal{C}_i} \rho_k-{p}_i\right| \leq \delta, i \in \mathcal{A}$, where $\rho_k=0.5(1+\cos\pi\lVert{p}_k-{p}_i\rVert / r_{\text{sense}})$, $\mathcal{C}_i$ is the sets
of neighbors/observed cells and $\delta=0.05$. 

\textit{\textbf{MARL Algorithm and Training:}}
We adopt the MADDPG algorithm \cite{lowe_multi-agent_2017} for the shape assembly
task due to its compatibility with continuous action domains and its off-policy characteristics. The underlying architecture follows an actor-critic design, where both networks are instantiated as Multilayer Perceptrons (MLPs). We employ Leaky-ReLU activation functions for the hidden layers, while the actor network's output is normalized using a Tanh function. The training parameters are shown as follows: The number of episodes = $3000$,
episode length = $200$, batch size = $512$, hidden dim = $180$,
number of hidden layers = $3$, lr-critic = $1$e$-3$, lr-actor = $1$e$-4$,
exploration rate = $0.6$, noise scale = $0.1$, gamma = $0.99$. To further bolster the policy's adaptability and robustness, we dynamically randomize the target shape throughout the training phase rather than relying on a static geometric configuration.

\subsection{Competitive Predator-prey Pursuit-evasion}

The predator-prey environment becomes more complex since it contains heterogeneous agents and dynamically-changed environment conditions.  To capture the diversity of real-world scenarios, two different environmental settings are considered. The first is a confined square area, where agents cannot cross the boundaries modeled as walls with a specified contact stiffness, as commonly used in prior studies \cite{lowe_multi-agent_2017}. The second is an unbounded area, designed to approximate vast, open-field domains in nature.  We employ a toroidal topology to simulate such infinite regions, where agents reappear on the opposite side after exiting the environment. This unbounded condition is widely employed in swarm modeling \cite{vicsek_novel_1995, durve_learning_2020} to approximate large or infinite domains.
\begin{figure}[h]
	\centering 
	\includegraphics[width=1\linewidth]{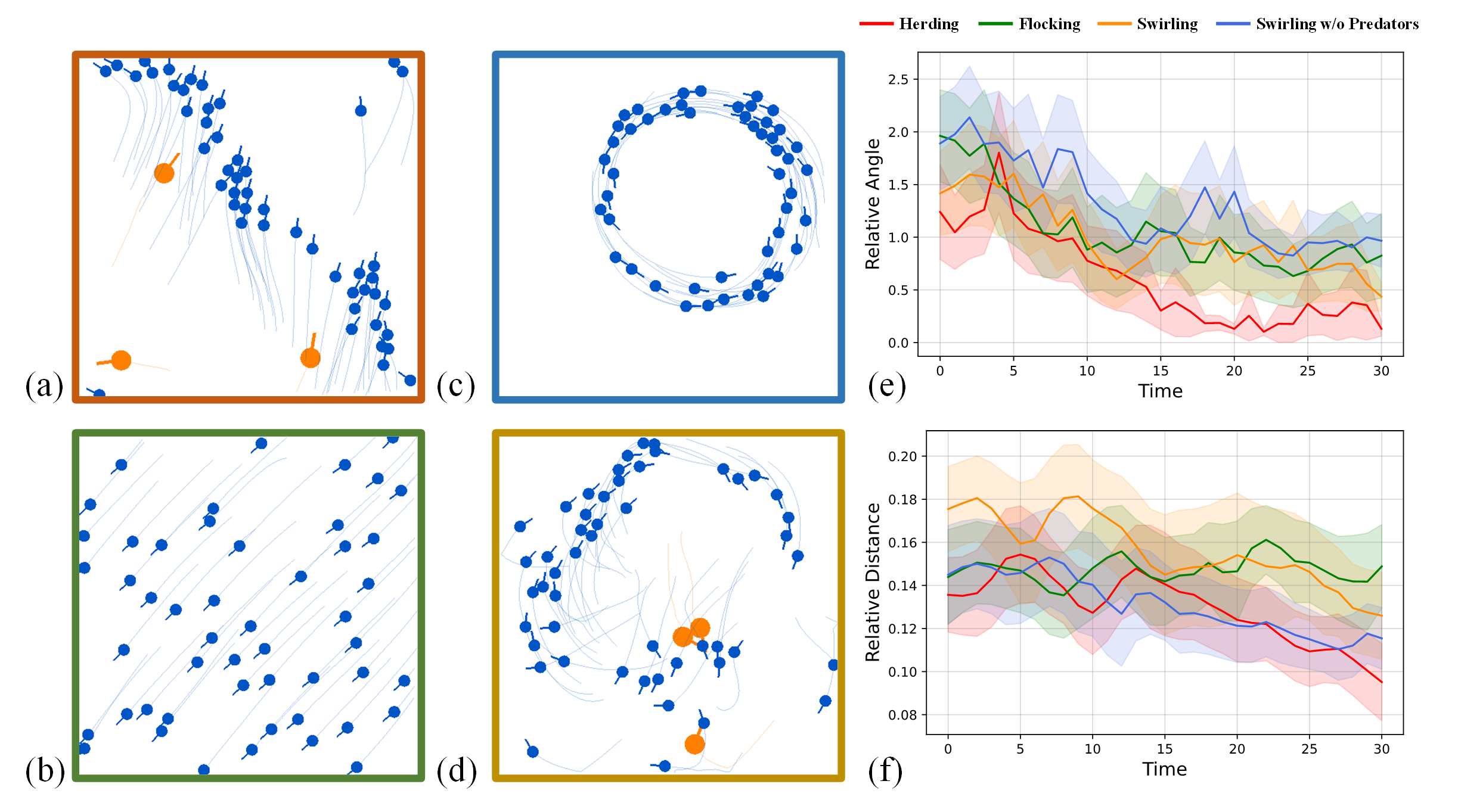}
	\caption{Heterogeneous agents: Orange circles represent predators and blue circles represent prey. Each agent is equipped with a short line segment indicating its orientation. (a) Herding. (b) Flocking. (c) Swirling without predators. (d) Swirling in the presence of predators.  (e) Change in relative orientation angle between each prey and its nearest prey. The shaded area indicates a $95\%$ confidence interval. (f) Change in relative distance. }\label{fig:case_theta_agree}
\end{figure}

After MARL training, prey exhibit distinct collective behaviors across scenarios. \cref{fig:case_theta_agree}(a)-(d) presents four scenarios where collective behaviors emerge. Specifically, (a) and (b) correspond to unbounded conditions, while (c) and (d) correspond to confined environments. In \cref{fig:case_theta_agree}(a), prey learned herding behavior in the presence of predators, characterized by two key features: orientations alignment and distance reduction, shown in \cref{fig:case_theta_agree}(e) and (f). \cref{fig:case_theta_agree}(b) shows flocking behavior in the absence of predators, where prey maintain orientation alignment but without a further decrease in distance. \cref{fig:case_theta_agree}(c) and (d) demonstrate the emergence of swirling behavior in the confined environment.

\begin{figure*}[h]
	\centering 
	\includegraphics[width=1\linewidth]{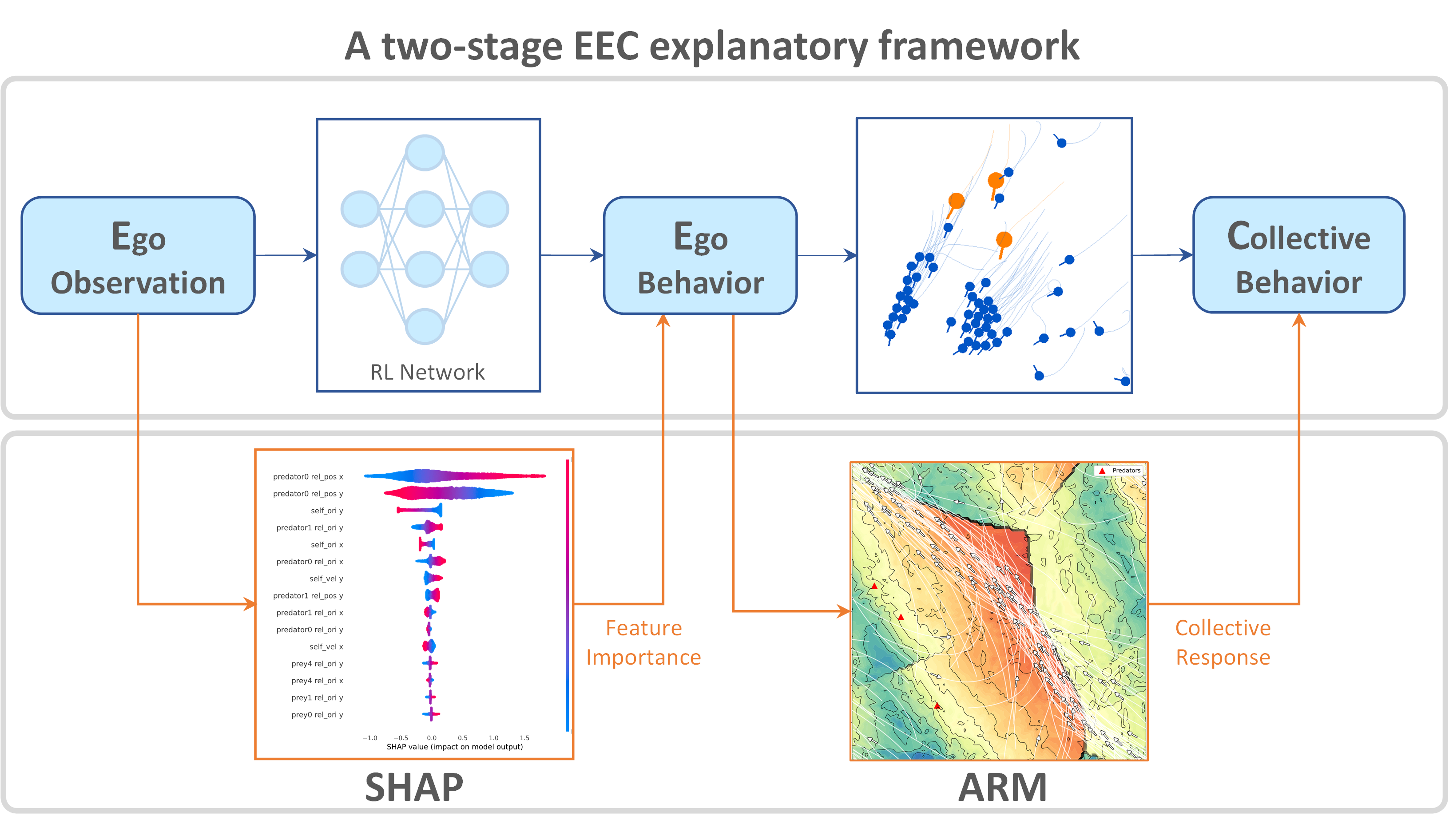}
	\caption{The proposed EEC explanation scheme comprises two components: 1) \LinkI, which reveals how ego-behavior depends on ego-observation by analyzing the influence of individual features in the observation vector and identifying the most influential features; 2) \LinkII, which uses ARM to analyze how these feature-driven behaviors give rise to collective phenomena. An example of the ARM  of $a_{\mathrm{R}}$ in predator-prey environment under unbounded condition is shown in bottom right, where red triangles denote predators, white lines depict real prey trajectories, and arrows indicate movement direction. Black contour lines show the spatial level sets of $a_{\mathrm{R}}$.
	}\label{fig:structure}
\end{figure*}The dynamics of agent motion can be described as:
\begin{equation}
	\begin{aligned}
		\theta(t+1) & =\theta(t)+a_{\mathrm{R}} \Delta t \\
		v(t+1) & =v(t)+\left(a_{\mathrm{F}} h+f_{\mathrm{d}}+f_{\mathrm{e}}\right) \Delta t / m_i \\
		x(t+1) & =x(t)+v(t) \Delta t 
	\end{aligned}\label{dynamics}
\end{equation}where $\Delta t$ represents the time step, $\theta \in (-\pi, \pi] \subset \mathbb{R}$ is the orientation angle. $a_R \in \mathbb{R}$ and $a_F \in \mathbb{R}$ are the two actions that an agent can perform, representing angular velocity and acceleration, respectively. $h = [\cos \theta, \sin \theta]^{\mathrm{T}} \in \mathbb{R}^2$ represents the agent's orientation. $f_{\mathrm{d}} \in \mathbb{R}^2$ and $f_{\mathrm{e}} \in \mathbb{R}^2$ are the drag force and elastic force caused by collisions between agents, respectively. $x \in \mathbb{R}^2$ represents the position  of the agent, and $v \in \mathbb{R}^2$ represents the velocity, while $m_i \in \mathbb{R}^{+}$ represents the mass. 

The observation vector for each agent is as follows:
\begin{equation}
	\left[\begin{array}{c}
		\text {Agent's own position, velocity, and orientation} \\
		\text {Relative position and orientation of observed predator} \\
		\text {Relative position and orientation of observed prey}
	\end{array}\right] \label{obs_vector}
\end{equation}The observation vector encodes the relative position and orientation of the nearest 6 prey and 6 predators within an omnidirectional field of view, ordered by proximity. This fixed-range, partially observable setting aligns with standard conventions in the MARL domain \cite{yu2022surprising} and requires agents to actively prioritize spatially distributed risks. Zero-padding is applied when fewer neighbors are detected. 

Rewards are grounded in survival instincts: predators gain $r_{\text{predator}} = +1$ and prey incur $r_{\text{prey}} = -1$ upon physical collision.  This reward mechanism
operates independently of the emergent collective behavior and does not directly incentivize grouping.
Despite using only this simple reward function, complex collective behaviors emerge. Additionally, we introduce
an energy-consumption penalty $-0.01\lvert a_F\rvert - 0.1\lvert a_R\rvert$, to better reflect the cost of movement in nature.  In the bounded environment setting, an additional penalty -0.1 is added to the reward function when contact between agents and boundaries happens. This setting is designed to simulate
either the presence of danger in the outside world.

{Collective behaviors emerge under various actor-critic MARL frameworks, such as MAPPO \cite{kolle2024aquarium} and MADDPG \cite{li_predatorprey_2023}. Both actor and critic networks employ 3-layer feed-forward architectures (64 units, ReLU). The actor outputs a 2-D action vector, while the critic estimates a scalar value.}

\begin{figure*}[h]
	\centering 
	\includegraphics[width=1\linewidth]{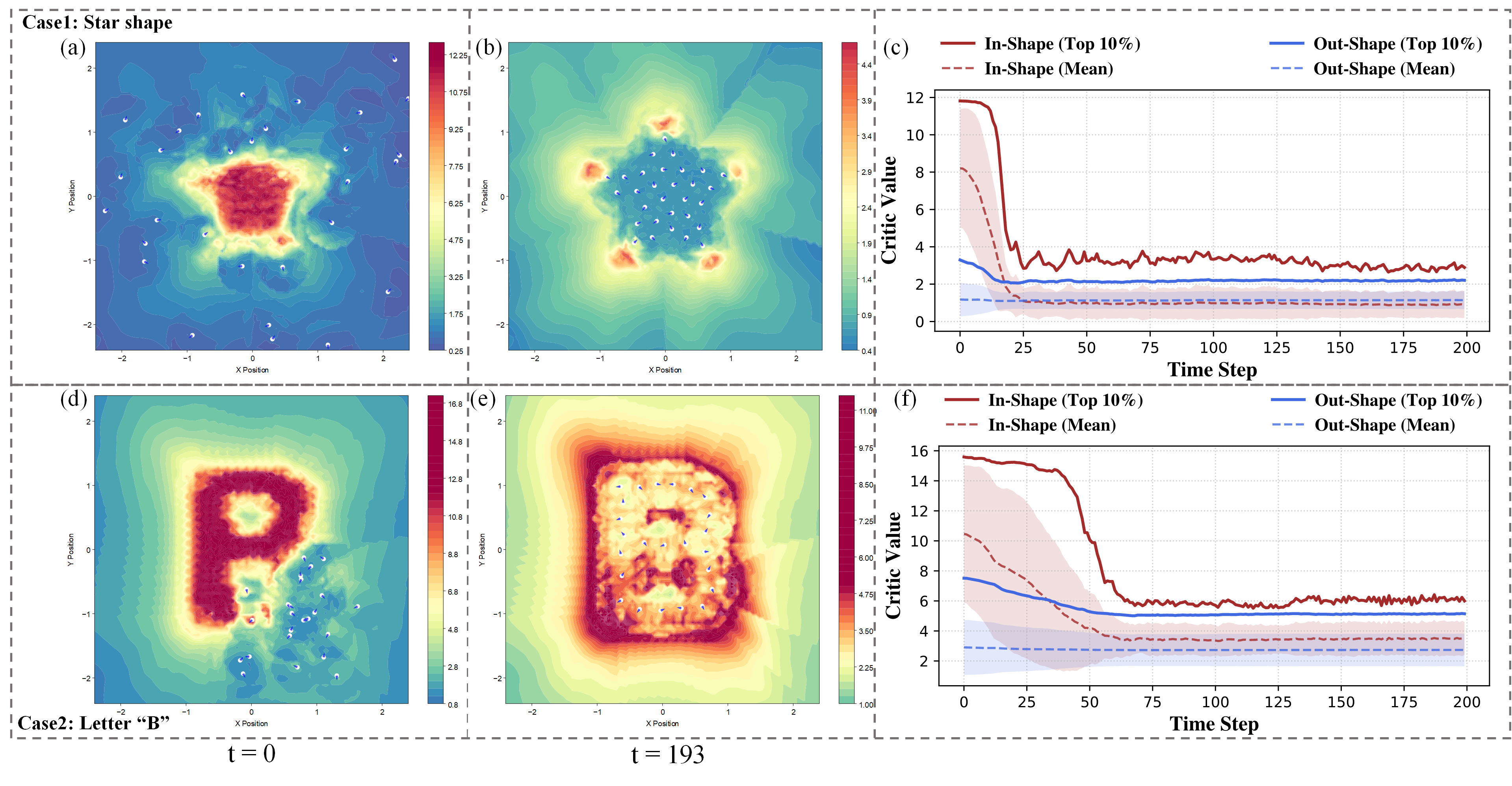}
	\caption{ARM for the shape-assembly task with two distinct target shapes: (a)-(c) show the results for the star shape, where (a)-(b) display the ARM of the critic network at different time steps. Color intensity indicates the magnitude of critic value. White circles represent robots. (c) Comparison of the overall mean and the top-$10\%$ mean of critic values within In-Shape versus Out-Shape regions. The shaded area indicates a $95\%$ confidence interval. (d)-(f) present the results for the letter ``B'' shape.
	}  
	\label{fig:ARM_shape}
\end{figure*}
\section{EEC Explanation Framework}\label{sec:explan}

Our goal is to reveal the mechanisms underlying the emergent multi-agent RL policies across these distinct tasks. The proposed EEC explanatory framework aims to uncover the mechanisms behind emergent swarming by exposing two hidden links from ego observation to global coordination, as illustrated in \cref{fig:structure}.

To decode the ego-observation to ego-behavior mapping (\LinkI{}), we employ SHAP \cite{lundberg_unified_2017} to quantify feature influence. SHAP can provide additive, signed, and comparable attributions across heterogeneous observation features. To distinguish causal necessity from correlation, we conduct ablation studies by retraining policies with masked features: Features whose removal collapses collective behavior are deemed critical, shown in \cref{exp:shap}.

To reveal how ego-behavior scales to collective behavior, we propose ARM, a counterfactual spatial policy operator. ARM probes policy under counterfactual spatial intervention and thus, can explain how local decisions organize into swarm patterns. This is achieved by introducing a virtual agent which is invisible to real agents and therefore does not influence their behavior. The virtual agent acts as a probe to query the prey's response at different spatial locations at the current timestep, which can avoid the non-stationarity issue caused by mutual interference among agents.  In this way, we can explain how key features shape agents movement, leading to collective behaviors. 

An example of ARM is shown in \cref{fig:structure}. The color of each pixel represents the policy output for the virtual prey's angular velocity $a_{\mathrm{R}}$ when it is placed at that location and its induced observation is fed into the policy network. We make three observations: 1) We observe that prey, initially scattered, gradually converge to a fixed region. This region is automatically highlighted by multiple black polylines through ARM. In fact, this region corresponds to the boundary of the predators' Voronoi diagram \cite{okabe2009spatial}. It is the desired target for coordinated motion: When a prey leaves this region, it quickly converges back. 
 2) The colors on the two sides of the Voronoi boundary are different, indicating distinct response regimes; 3) The color distribution depends on predator locations, i.e., the prey adjusts $a_{\mathrm{R}}$ according to its relative position to the predators.

The convergence of prey towards the Voronoi boundary remains robust under different agent parameter settings, e.g., different populations and speed ratios. We also analyze collective behavior in confined environments to verify that the agents' policy can adapt to boundary constraints. Detailed analyses of these findings are presented in \cref{heading_consistent}.

In the following, we aim to analyze the mechanisms underlying the emergence of  the collective behavior in both cooperative and competitive task.  

\begin{figure*}[h]
	\centering 
	\includegraphics[width=1\linewidth]{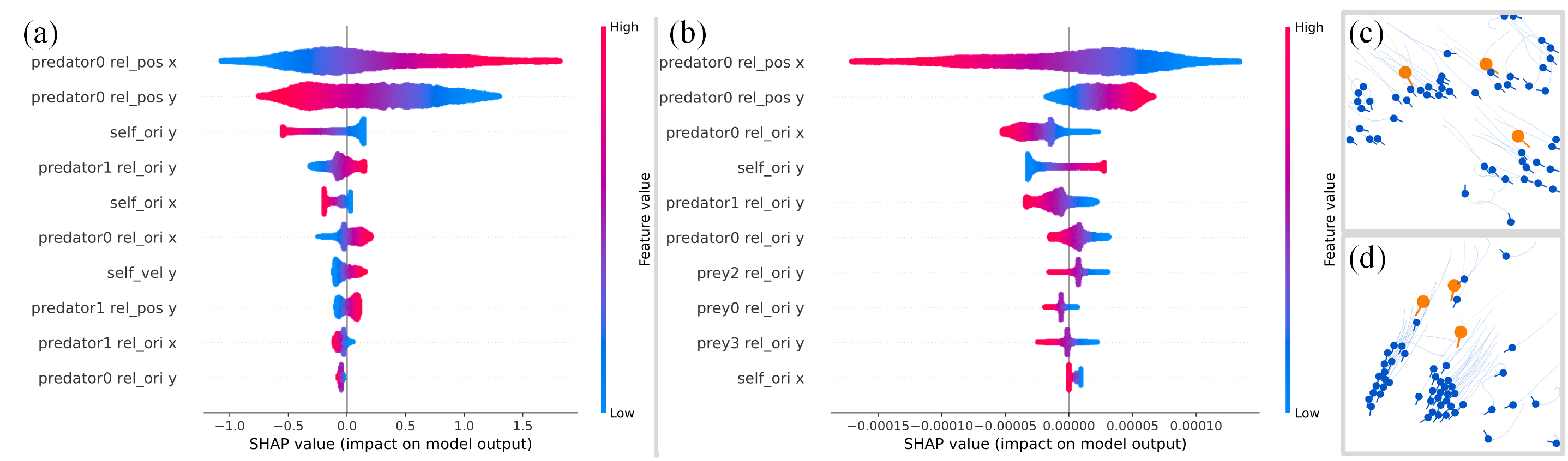}
	\caption{(a) SHAP analysis shows the top 10 observation features influencing $a_{\mathrm{R}}$. The $x$-axis represents SHAP value, while the $y$-axis lists the corresponding features. Positive SHAP values indicate positive contributions to the action, negative values indicate inhibitory effects, and larger magnitudes denote stronger influence.  (b) SHAP analysis of $a_{\mathrm{F}}$. (c) Ablation: Removing orientation features. (d) Ablation: Fixing acceleration at maximum.}\label{fig:shap_value_herding}
\end{figure*}
\section{Analysis of Cooperative Task}
For the cooperative shape assembly task, the primary challenge is balancing goal-reaching constraint with collision avoidance. To explain how the learned policy manages these spatial constraints, we employ the ARM to reveal variation of the robot's critic value in the task space. This analysis uncovers the temporal evolution of the target regions prioritized by the robots throughout the task.

We consider two distinct target shapes: a five-pointed star and the letter ``B". The letter ``B" represents a more challenging scenario characterized by non-convex and multi-component geometric properties. Furthermore, we conducted a noise perturbation on the letter ``B'' task by injecting Gaussian white noise ($\sigma=0.1$) into the robots' observation vectors  to evaluate the robustness of ARM. The results are shown in \cref{fig:ARM_shape}. ARM is constructed to reveal the robots' responses across different spatial locations: A virtual robot is introduced to record observation data throughout the environmental space, which is then fed into the trained RL networks. The resulting output is mapped onto the environment space to reveal the underlying mechanisms of the agents' behavior.
From \cref{fig:ARM_shape}(a)-(b), we observe that initially, the unoccupied target center exhibits high critic values, drawing robots inward. As the center fills, the value peak shifts to the boundary, promoting the exploration of unoccupied areas to mitigate overcrowding within the shape's interior. This mirrors the design intuition in  \cite{sun2023mean}. We support this finding with quantitative analysis in \cref{fig:ARM_shape}(c), comparing the overall mean and top $10\%$ mean critic values inside versus outside the target shape. Initially, internal values far exceed external ones. As the task progresses, the internal mean value drops due to robots occupancy, while external values remain consistently low. The In-Shape Top $10\%$ value remains higher than the Out-Shape. 

For the letter ``B" shape assembly task, although the shape becomes more complex with additional noise, the results are similar, as shown in \cref{fig:ARM_shape}(d)-(f). High spatial values initially guide robots to fill unoccupied regions, subsequently shifting to the boundaries upon completion to promote outward exploration and mitigate internal overcrowding.

These results validate the effectiveness of ARM in identifying meaningful geometric invariants in cooperative MARL.

\section{Analysis of Competitive Task}
In the competitive pursuit-evasion task, the agent faces a more complex, high-dimensional observation space involving adversaries and peers. Unlike the static desired targets in shape assembly task, the dynamic nature of the pursuit-evasion requires us to first identify which observation features drive the decision-making process. Therefore, we first apply SHAP to identify the influential features in the observation vector to reveal \LinkI. Then  we uses ARM to analyze how these feature-driven behaviors in the environment give rise to collective phenomena, unveiling \LinkII. We begin our analysis with the unbounded space scenario, where the absence of obstacles allows us to isolate the evasion condition of prey agents. Subsequently, we extend our study to confined environments to verify whether the agents have learned to avoid collisions while evading.
\begin{figure*}[h]
	\centering 
	\includegraphics[width=1\linewidth]{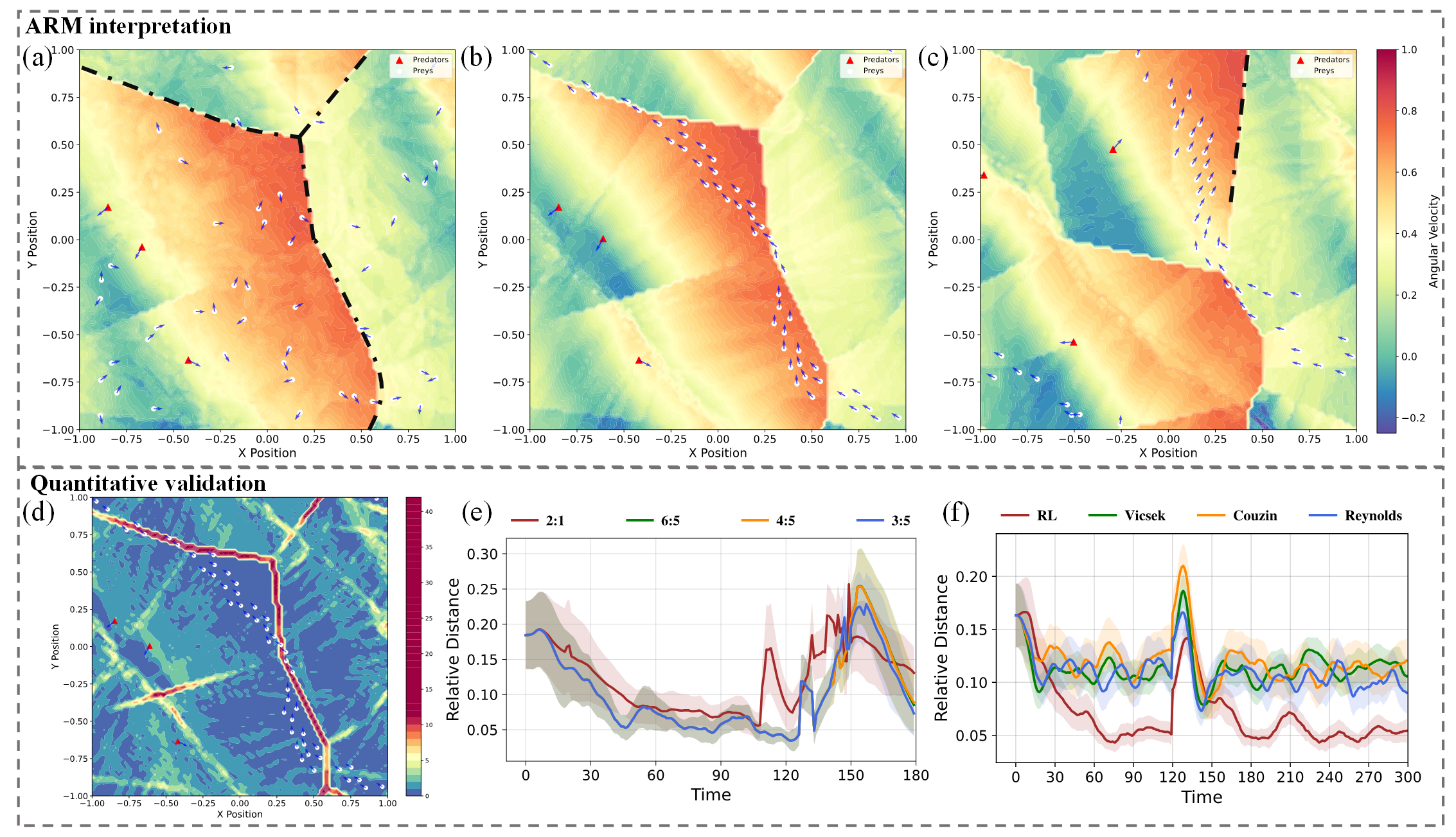}
	\caption{ARM analysis for pursuit-evasion task: (a)-(c) show the ARM of $a_{\mathrm{R}}$ at three different time steps. Color intensity indicates the output magnitude. Red triangles denote predators, white circles represent prey, and arrows indicate orientation. (d) Spatial Gradient Magnitude of ARM. Color intensity indicates the magnitude of gradient. (e) The prey's relative distance to the boundary of the predators' Voronoi diagram under four different $v_{\text {predator }}: v_{\text {prey }}$speed ratios. The shaded area indicates a $95\%$ confidence interval. (f) The RL results compared with three rule-based baselines \cite{vicsek_novel_1995, reynolds1987flocks, couzin_collective_2002}.
	}\label{fig:ARM_herding}
\end{figure*}
\subsection{From Ego-Observation to Ego-Behavior}\label{exp:shap}
To assess feature contributions to the policy network, we employ SHAP \cite{lundberg_unified_2017}. Unlike standard i.i.d. applications, we apply SHAP to a multi-agent, partially observable, continuous-control setting, attributing egocentric features (relative positions/orientations) to policy outputs.  The analysis uses observation data collected from 50 prey agents over 500 time steps, with 100 random background samples.

\textbf{\textit{Analysis of Observation Features Influencing Action $a_{\mathrm{R}}$: }}It is shown that the nearest predator's relative position has the strongest effect on $a_{\mathrm{R}}$, demonstrated in \cref{fig:shap_value_herding}(a). This suggests prey adjust orientation primarily based on predator proximity.

Despite an isotropic environment, the $x$ and $y$ components yield different SHAP values. This likely stems from the non-i.i.d. nature of RL data inducing directional bias and the context-dependent nature of SHAP attributions (e.g., identical ranges require different turns depending on bearing). These limitations motivate our ARM method (Sec.~\ref{heading_consistent}).

{To verify that the feature importance identified by SHAP reflects causality rather than only correlation, we conducted causal feature ablation studies by retraining the policy with specific features masked and tested on 20 different random seeds.} First, removing neighbor orientation features did not hinder alignment or collective behavior (\cref{fig:shap_value_herding}(c)). Conversely, removing relative position features completely eliminated collective coordination. This causal intervention confirms that relative position, not orientation, is the critical feature for collective coordination.

\textit{\textbf{Analysis of Observation Features Influencing Action $a_{\mathrm{F}}$: }}Counterintuitively, $a_{\mathrm{F}}$ shows weak correlation with observation features (\cref{fig:shap_value_herding}(b)). Even the predator's position, despite being the top contributor, has a negligible SHAP value, indicating limited environmental influence on $a_{\mathrm{F}}$. 

To validate the causal role of acceleration, we fixed $a_{\mathrm{F}}$ to its maximum value during execution. As shown in \cref{fig:shap_value_herding}(d), prey still formed compact groups, confirming that herding is not driven by $a_{\mathrm{F}}$.

\textit{\textbf{Summation: }}Taken together, the SHAP and ablation analyses show that the relative position of the nearest predator is the dominant factor influencing $a_{\mathrm{R}}$, while $a_{\mathrm{F}}$ exhibits limited sensitivity to observations. We thus focus next on how predator position shapes prey decisions via $a_{\mathrm{R}}$.

\subsection{From Ego-Behavior to Collective Behavior}\label{heading_consistent}
In this section, we introduce ARM to show how ego-behavior gives rise to collective behavior. Due to continuous interactions among agents, the environment evolves dynamically, making consistent behavioral patterns hard to isolate. To control for this, we keep the predator stationary from $t=0$ to $t=100$, then let it act under its learned policy, and fix it again for $t\geq 150$. This setup allows us to isolate and analyze the prey's behavioral responses under stable and transitional conditions.

To reveal how the policy and value functions vary across space, we construct the ARM: A virtual prey agent is introduced to record observed data across the environment space, which is then fed into the trained RL networks. Under the ARM interpretation, it is shown that the prey's objective is to reach and move along the boundary of the predators' Voronoi diagram, as demonstrated in \cref{fig:ARM_herding}(a)-(c). It presents the ARM interpretation results at three representative times: initialization $t=0$, stationary predators $t=80$, and predator pursuit $t=146$. When the predators move, the prey track this shifting boundary. The ARM automatically reveals an obvious discontinuity at the Voronoi boundary. This arises because prey on opposite sides of the boundary must rotate in divergent directions to evade the predator; this sharp contrast in angular velocity values creates the observed discontinuity.

The ARM results are validated by two quantitative approaches. First, we introduce the Spatial Gradient Magnitude (SGM) of ARM to quantify discontinuities by calculating the gradient magnitude of ARM outputs with respect to spatial position. Higher values indicate sharper action transitions. As shown in \cref{fig:ARM_herding}(d), SGM peaks near Voronoi boundaries (magnitude $\approx 40$) while remaining below $5$ elsewhere, confirming the sharp transitions identified by ARM.

Second, to verify the Voronoi boundary as the convergence target, we track prey-boundary distance under two conditions (\cref{fig:ARM_herding}(e, f)): dynamic predator movement in \cref{fig:ARM_herding}(e) ($100 \le t \le 150$)  and  impulse perturbations in \cref{fig:ARM_herding}(f) ($t=120$):
 1) In (e), we evaluate prey reconvergence under dynamic predator movement  across four speed ratios. The distance stabilizes near $0.05$, fluctuates during motion ($t=100$), and recovers within $30$s after stopping ($t \ge 150$). Higher speed ratios ($2:1$) induce larger deviations and slower recovery.
2) In (f), compared to three baselines \cite{vicsek_novel_1995, reynolds1987flocks, couzin_collective_2002}. The RL model achieves a steady-state distance of $0.05$, outperforming the baselines ($\approx 0.10$). Upon applying an impulse perturbation at $t=120$, the deviation distance of the RL model remains below $0.15$ and rapidly reconverges, whereas baseline deviations all exceed $0.15$. This stability analysis confirms that the Voronoi boundary is the prey's desired convergence target and that the RL method outperforms typical swarm models in terms of convergence speed.


\textit{\textbf{Analysis of Collective Behavior in Confined Environment:}}
Real-world robotic deployments mainly operate within confined environments. Therefore, verifying that the agent's policy can adapt to boundary constraints while maintaining cohesion is a critical step towards sim-to-real transfer.

However, analyzing the policy in confined area is challenging because the prey's behavior is driven by coupled dynamics: the repulsive force from the predator and the collision penalties from the walls. It is difficult to distinguish whether a turn is caused by the predator's direction or the wall's proximity. To decouple these factors and strictly verify the boundary-avoidance mechanism, we devise a controlled  experiment: Instead of a dynamic predator, we override the prey's observation of the predator's relative position to a fixed value of $(0,0)$, which means each prey perceives the nearest predator to be located exactly at its own current position. This removes all directional cues from the adversary, and allows us to attribute resulting spatial structure solely to the environmental boundaries.

\begin{figure}[h]
	\centering 
	\includegraphics[width=\linewidth]{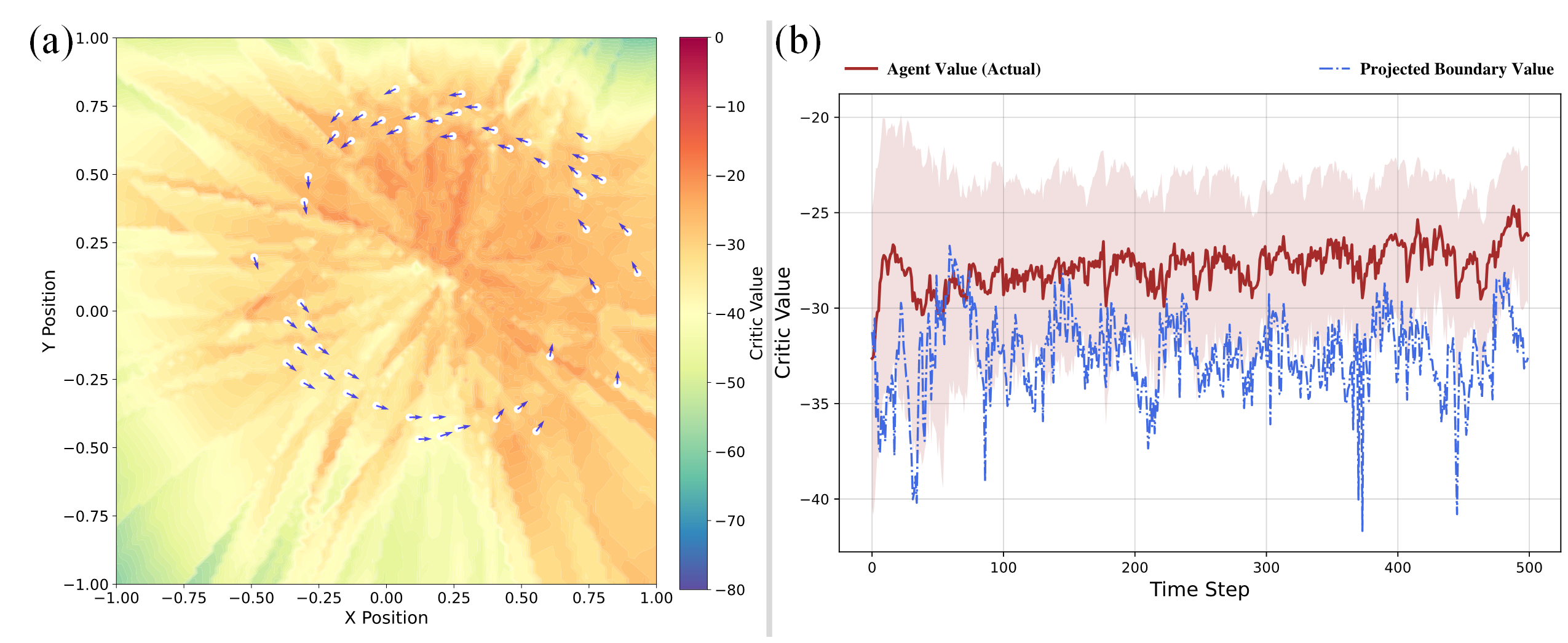}
	\caption{ARM Analysis for confined environment: (a) ARM of the critic network. (b) Quantitative verification of obstacle avoidance. The red curve shows the average critic value of prey. The blue dashed curve represents the counterfactual value if agents were projected to the nearest boundary.
	}\label{fig:ARM_bounded_space}
\end{figure}

Under this controlled setting, we observe that the agents do not crash into walls, nor do they move randomly. Instead, they spontaneously generate a stable swirling trajectory, shown in \cref{fig:ARM_bounded_space}(a). The ARM analysis reveals that the critic value diminishes as the agent approaches the boundaries. The quantitative analysis shown in \cref{fig:ARM_bounded_space}(b) further demonstrates this finding: The agent's actual critic value is higher than that of a hypothetical agent projected onto the nearest boundary. This indicates that the observed swirling represents a desired geometric strategy in confined area for wall avoidance.

\section{Conclusion}
This paper proposed ARM to explore the hidden mechanisms of collective behavior trained through MARL. ARM reveals agents implicitly learn geometric structure and use it as desired targets for coordinated motion. These findings offer new insights into collective behavior in MARL.

\bibliography{nips25}

\bibliographystyle{IEEEtran}

\end{document}